\documentclass[11pt,a4paper]{article}
\usepackage[hyperref]{emnlp-ijcnlp-2019}
\usepackage{times}
\usepackage{enumitem}
\usepackage{latexsym}

\usepackage{url}
\usepackage{times}
\usepackage{latexsym}
\usepackage{colortbl}
\usepackage{tikz}
\usepackage{amsmath}
\usepackage{breqn}
\usetikzlibrary{shapes.geometric, arrows}
\usepackage{tikz}
\usetikzlibrary{matrix}
\usepackage{lipsum,adjustbox}
\usetikzlibrary{calc,positioning}
\usepackage{graphicx} 
\usepackage{siunitx}
\usepackage{amsmath,amssymb}
\aclfinalcopy 

\tikzset{ 
    table/.style={
        matrix of nodes,
        row sep=-\pgflinewidth,
        column sep=-\pgflinewidth,
        nodes={
            rectangle,
            draw=black,
            align=center
        },
        minimum height=1.5em,
        text depth=0.5ex,
        text height=2ex,
        nodes in empty cells,
        every even row/.style={
            nodes={fill=white!20}
        },
        column 1/.style={
            nodes={text}
        },
        row 1/.style={
            nodes={
                fill=white,
                text=black
            }
        }
    }
}

\tikzstyle{target_layer} = [rectangle,  text centered, draw=black, fill=white!10]

\tikzstyle{target_e_layer} = [rectangle,  text centered, draw=black, fill=green!30]

\tikzstyle{last_layer} = [rectangle,  text centered, draw=black, fill=yellow!30]

\tikzstyle{middle_layer} = [rectangle,  text centered, draw=black, fill=white!30]

\tikzstyle{first_layer} = [rectangle,   text centered, draw=black, fill=orange!30]

\tikzstyle{cnn_layer} = [rectangle, text centered, draw=black, fill=white!30]
\tikzstyle{bin_layer} = [rectangle, text centered, draw=black, fill=white!30]
\tikzstyle{score_layer} = [rectangle, text centered, draw=white, fill=white!30]

\tikzstyle{concat_layer} = [rectangle, text centered, draw=white, fill=white!30]

\tikzstyle{arrow} = [thick,->,>=stealth]

\title{Entity-aware ELMo: Learning Contextual Entity Representation for Entity Disambiguation}

\author{First Author \\
  Affiliation / Address line 1 \\
  Affiliation / Address line 2 \\
  Affiliation / Address line 3 \\
  {\tt email@domain} \\\And
  Second Author \\
  Affiliation / Address line 1 \\
  Affiliation / Address line 2 \\
  Affiliation / Address line 3 \\
  {\tt email@domain} \\}

\author{Hamed Shahbazi, Xiaoli Z. Fern, Reza Ghaeini, \\\textbf{Rasha Obeidat, Prasad Tadepalli}
\\
Oregon State University, Corvallis, OR, USA \\
\{shahbazh, xfern, ghaeinim, obeidatr, tadepall\}@eecs.oregonstate.edu
}

\date{}

\begin{document}
\maketitle
\begin{abstract}
  We present a new local entity disambiguation system. The key to our system is a novel approach for learning entity representations. In our approach we learn an entity aware extension of Embedding for Language Model (ELMo) which we call Entity-ELMo (E-ELMo). Given a paragraph containing one or more named entity mentions, each mention is first defined as a function of the entire paragraph (including other mentions), then they predict the referent entities. Utilizing E-ELMo for local entity disambiguation, we outperform all of the state-of-the-art local and global models on the popular benchmarks by improving about 0.5\% on micro average accuracy for AIDA test-b with Yago candidate set. The evaluation setup of the training data and candidate set are the same as our baselines for fair comparison.
\end{abstract}

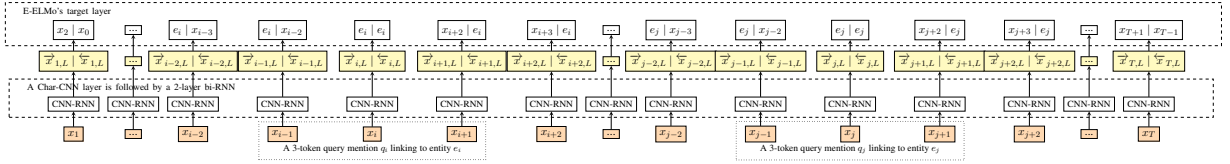
\begin {figure*}[ht]
\centering
\normalsize
\begin{adjustbox}{width=\textwidth}
\begin{tikzpicture}

\node (fl1) [first_layer] {\normalsize $x_1$};
\node (fl2) [first_layer, right of=fl1, xshift=1cm] {...};
\node (fl3) [first_layer, right of=fl2, xshift=1cm] {\normalsize $x_{i-2}$};
\node (fl4) [first_layer, right of=fl3, xshift=2cm] {\normalsize $x_{i-1}$};
\node (fl5) [first_layer, right of=fl4, xshift=2cm] {\normalsize $x_{i}$};
\node (fl6) [first_layer, right of=fl5, xshift=2cm] {\normalsize $x_{i+1}$};
\node (fl7) [first_layer, right of=fl6, xshift=2cm] {\normalsize $x_{i+2}$};
\node (fl8) [first_layer, right of=fl7, xshift=1cm] {...};
\node (fl9) [first_layer, right of=fl8, xshift=1cm] {\normalsize $x_{j-2}$};
\node (fl10) [first_layer, right of=fl9, xshift=2cm] {\normalsize $x_{j-1}$};
\node (fl11) [first_layer, right of=fl10, xshift=2cm] {\normalsize $x_{j}$};
\node (fl12) [first_layer, right of=fl11, xshift=2cm] {\normalsize $x_{j+1}$};
\node (fl13) [first_layer, right of=fl12, xshift=2cm] {\normalsize $x_{j+2}$};
\node (fl14) [first_layer, right of=fl13, xshift=1cm] {...};
\node (fl15) [first_layer, right of=fl14, xshift=1cm] {\normalsize $x_T$};

\node (cl1) [cnn_layer, above of=fl1, yshift=0.01cm] {\small CNN-RNN};
\node (cl2) [cnn_layer, right of=cl1, xshift=1cm] {\small CNN-RNN};
\node (cl3) [cnn_layer, right of=cl2, xshift=1cm] {\small CNN-RNN};
\node (cl4) [cnn_layer, right of=cl3, xshift=2cm] {\small CNN-RNN};
\node (cl5) [cnn_layer, right of=cl4, xshift=2cm] {\small CNN-RNN};
\node (cl6) [cnn_layer, right of=cl5, xshift=2cm] {\small CNN-RNN};
\node (cl7) [cnn_layer, right of=cl6, xshift=2cm] {\small CNN-RNN};
\node (cl8) [cnn_layer, right of=cl7, xshift=1cm] {\small CNN-RNN};
\node (cl9) [cnn_layer, right of=cl8, xshift=1cm] {\small CNN-RNN};
\node (cl10) [cnn_layer, right of=cl9, xshift=2cm] {\small CNN-RNN};
\node (cl11) [cnn_layer, right of=cl10, xshift=2cm] {\small CNN-RNN};
\node (cl12) [cnn_layer, right of=cl11, xshift=2cm] {\small CNN-RNN};
\node (cl13) [cnn_layer, right of=cl12, xshift=2cm] {\small CNN-RNN};
\node (cl14) [cnn_layer, right of=cl13, xshift=1cm] {\small CNN-RNN};
\node (cl15) [cnn_layer, right of=cl14, xshift=1cm] {\small CNN-RNN};

\node (ll1) [last_layer, above of=cl1, yshift=.45cm] {\normalsize $\overrightarrow{x}_{1,L} \mid \overleftarrow{x}_{1,L}$};
\node (ll2) [last_layer, right of=ll1, xshift=1cm] {...};
\node (ll3) [last_layer, right of=ll2, xshift=1cm] {\normalsize $\overrightarrow{x}_{i-2,L} \mid \overleftarrow{x}_{i-2,L}$};
\node (ll4) [last_layer, right of=ll3, xshift=2cm] {\normalsize $\overrightarrow{x}_{i-1,L}\mid \overleftarrow{x}_{i-1,L}$};
\node (ll5) [last_layer, right of=ll4, xshift=2cm] {\normalsize $\overrightarrow{x}_{i,L}\mid \overleftarrow{x}_{i,L}$};
\node (ll6) [last_layer, right of=ll5, xshift=2cm] {\normalsize $\overrightarrow{x}_{i+1,L}\mid \overleftarrow{x}_{i+1,L}$};
\node (ll7) [last_layer, right of=ll6, xshift=2cm] {\normalsize $\overrightarrow{x}_{i+2,L}\mid \overleftarrow{x}_{i+2,L}$};
\node (ll8) [last_layer, right of=ll7, xshift=1cm] {...};
\node (ll9) [last_layer, right of=ll8, xshift=1cm] {\normalsize $\overrightarrow{x}_{j-2,L} \mid \overleftarrow{x}_{j-2,L}$};
\node (ll10) [last_layer, right of=ll9, xshift=2cm] {\normalsize $\overrightarrow{x}_{j-1,L}\mid \overleftarrow{x}_{j-1,L}$};
\node (ll11) [last_layer, right of=ll10, xshift=2cm] {\normalsize $\overrightarrow{x}_{j,L}\mid \overleftarrow{x}_{j,L}$};
\node (ll12) [last_layer, right of=ll11, xshift=2cm] {\normalsize $\overrightarrow{x}_{j+1,L}\mid \overleftarrow{x}_{j+1,L}$};
\node (ll13) [last_layer, right of=ll12, xshift=2cm] {\normalsize $\overrightarrow{x}_{j+2,L}\mid \overleftarrow{x}_{j+2,L}$};
\node (ll14) [last_layer, right of=ll13, xshift=1cm] {...};
\node (ll15) [last_layer, right of=ll14, xshift=1cm] {\normalsize $\overrightarrow{x}_{T,L}\mid \overleftarrow{x}_{T,L}$};

\node (tl1) [target_layer, above of=ll1, yshift=0.05cm] {$x_{2}\mid x_{0}$};
\node (tl2) [target_layer, right of=tl1, xshift=1cm] {...};
\node (tl3) [target_layer, right of=tl2, xshift=1cm] {$e_{i}\mid x_{i-3}$};
\node (tl4) [target_layer, right of=tl3, xshift=2cm] {$e_{i}\mid x_{i-2}$};
\node (tl5) [target_layer, right of=tl4, xshift=2cm] {$e_{i}\mid e_{i}$};
\node (tl6) [target_layer, right of=tl5, xshift=2cm] {$x_{i+2}\mid e_{i}$};
\node (tl7) [target_layer, right of=tl6, xshift=2cm] {$x_{i+3}\mid e_{i}$};
\node (tl8) [target_layer, right of=tl7, xshift=1cm] {...};
\node (tl9) [target_layer, right of=tl8, xshift=1cm] {$e_{j}\mid x_{j-3}$};
\node (tl10) [target_layer, right of=tl9, xshift=2cm] {$e_{j}\mid x_{j-2}$};
\node (tl11) [target_layer, right of=tl10, xshift=2cm] {$e_{j}\mid e_{j}$};
\node (tl12) [target_layer, right of=tl11, xshift=2cm] {$x_{j+2}\mid e_{j}$};
\node (tl13) [target_layer, right of=tl12, xshift=2cm] {$x_{j+3}\mid e_{j}$};
\node (tl14) [target_layer, right of=tl13, xshift=1cm] {...};
\node (tl15) [target_layer, right of=tl14, xshift=1cm] {$x_{T+1}\mid x_{T-1}$};

\draw [arrow] (fl1) -- (cl1);
\draw [arrow] (fl2) -- (cl2);
\draw [arrow] (fl3) -- (cl3);
\draw [arrow] (fl4) -- (cl4);
\draw [arrow] (fl5) -- (cl5);
\draw [arrow] (fl6) -- (cl6);
\draw [arrow] (fl7) -- (cl7);
\draw [arrow] (fl8) -- (cl8);
\draw [arrow] (fl9) -- (cl9);
\draw [arrow] (fl10) -- (cl10);
\draw [arrow] (fl11) -- (cl11);
\draw [arrow] (fl12) -- (cl12);
\draw [arrow] (fl13) -- (cl13);
\draw [arrow] (fl14) -- (cl14);
\draw [arrow] (fl15) -- (cl15);

\draw [arrow] (cl1) -- (ll1);
\draw [arrow] (cl2) -- (ll2);
\draw [arrow] (cl3) -- (ll3);
\draw [arrow] (cl4) -- (ll4);
\draw [arrow] (cl5) -- (ll5);
\draw [arrow] (cl6) -- (ll6);
\draw [arrow] (cl7) -- (ll7);
\draw [arrow] (cl8) -- (ll8);
\draw [arrow] (cl9) -- (ll9);
\draw [arrow] (cl10) -- (ll10);
\draw [arrow] (cl11) -- (ll11);
\draw [arrow] (cl12) -- (ll12);
\draw [arrow] (cl13) -- (ll13);
\draw [arrow] (cl14) -- (ll14);
\draw [arrow] (cl15) -- (ll15);

\draw [arrow] (ll1) -- (tl1);
\draw [arrow] (ll2) -- (tl2);
\draw [arrow] (ll3) -- (tl3);
\draw [arrow] (ll4) -- (tl4);
\draw [arrow] (ll5) -- (tl5);
\draw [arrow] (ll6) -- (tl6);
\draw [arrow] (ll7) -- (tl7);
\draw [arrow] (ll8) -- (tl8);
\draw [arrow] (ll9) -- (tl9);
\draw [arrow] (ll10) -- (tl10);
\draw [arrow] (ll11) -- (tl11);
\draw [arrow] (ll12) -- (tl12);
\draw [arrow] (ll13) -- (tl13);
\draw [arrow] (ll14) -- (tl14);
\draw [arrow] (ll15) -- (tl15);

\draw[black,thin,dotted] ($(fl4.north west)+(-0.3,0.15)$)  rectangle ($(fl6.south east)+(0.3,-0.6)$) node[midway,below] {\small A 3-token query mention $q_i$ linking to entity $e_i$}++(-2,0.3);

\draw[black,thin,dotted] ($(fl10.north west)+(-0.3,0.15)$)  rectangle ($(fl12.south east)+(0.3,-0.6)$) node[midway,below] {\small A 3-token query mention $q_j$ linking to entity $e_j$}++(-2,0.3);



\draw[black,thin, dashed] ($(cl1.north west)+(-1.3,0.6)$)  rectangle ($(cl15.south east)+(1.3,-0.2)$);

\node[text width=10cm] at ($(cl1.north west)+(4.3,.3)$) {\small A Char-CNN layer is followed by a 2-layer bi-RNN};

\draw[black,thin, dashed] ($(tl1.north west)+(-1.6,0.6)$)  rectangle ($(tl15.south east)+(1.3,-0.2)$);

\node[text width=6cm] at ($(tl1.north west)+(2.0,.3)$) {\small E-ELMo's target layer};

\end{tikzpicture}
\end{adjustbox}
\caption{Bidirectional language model predicts different targets in ELMo and E-ELMo.}
\vspace{-.8cm}
\label{fig:elmo}
\end{figure*}
 
\section{Introduction}
Named Entity Disambiguation (NED) is an essential task in natural language processing that resolves mentions in a document to their referent entities in a Knowledge Base (KB). 
A notable differentiating factor between NED systems is whether the global joint inference is used to resolve all the mentions in the same document collectively, separating the local and global NED models. Local NED systems disambiguate a mention individually by utilizing the local compatibility between the mention (and its textual context) and its candidate entities. Global NED models further consider the global coherence between assigned entities via structured prediction. Empirically, they have been shown to consistently outperform local models as they capture long-range document-wise information. Nevertheless, in this work, we demonstrate that a simple local model, when equipped with the right context and entity representation, can achieve competitive, even superior performance compared to the state-of-the-art global models. 

There are several limitations with the existing local models ~\citep{l:he, l:sun, l:zhiting, l:yamada, l:yammada17, rel:hoffman, rel:ibm, l:phongle} in capturing contextual dependencies between mention and candiadate entities. Yamada et al.~\citeyearpar{l:yamada,l:yammada17} and Ganea et al. \citeyearpar{rel:hoffman} learn entity representations by encoding the co-occurrence statistics between the entity and the words in its context. Each word is considered individually and hence its syntactic and semantic roles in the sentence are overlooked. In another effort, Sil et al.~\citeyearpar{rel:ibm} propose a model in which a recurrent neural network is utilized to compare the context of the query mention with the canonical pages of all candidate entities. To make their approach practical, their method is limited to only consider the first paragraph in each canonical page, which may not provide sufficient information for representing an entity throughout a large corpus.  
A convolutional architecture has also been explored for encoding sentence-level contextual information  ~\citep{l:sun}. However its fixed length windows limit the scope of the context. 


Recently, Embedding for Language Model (ELMo)~\citep{l:elmo} is introduced to produce context sensitive representation of words as a function of the entire sentence. Although ELMo produces context-sensitive representations for words in a sentence, its learning objective is unaware of the entities. For instance in the phrase \textit{``Jordan as a member of the Tar Heels' national championship team},'' the language model predicts ambiguous mention \textit{Jordan}, instead of the entity \textit{Michael Jordan}. 
There are many virtues in ELMo which make it a suitable choice to be used for learning entity representations: 1) Each token is represented as a function of its surrounding context via bi-directional RNNs which can potentially capture the dependencies between a mention and all the surrounding context as well as other named entity mentions in the context. 2) Deep layers of ELMo capture syntactic and semantic dependencies both of which are required for NED, and 3) ELMo is trainable on un-annotated corpora which is an important means to transfer information. 

In this work we introduce a novel approach for learning contextual entity representations by learning an entity-aware extension of ELMo, which is surprisingly effective. The learning mechanism which we call E-ELMo trains the language model to predict the grounded entity when encountering its mentions, as opposed to the words in the mentions. This modification affords us context-rich entity representations that are well-suited for disambiguation of the named entities with just local contexts. Incorporating E-ELMo's representations into a very simple local model, we achieve a superior or competitive performance on popular benchmarks compared to the state-of-the-art global models. 

\section{Entity-ELMo (E-ELMo)}
Here we first briefly review ELMo (Sec~\ref{sec:elmorev}) and then explain E-ELMo (Sec~\ref{sec:eelmorev}). As an example, we will consider an instance paragraph, which is a sequence of $T$ tokens containing two mentions with three tokens each, as shown in Figure~\ref{fig:elmo}. Here mentions $[x_{i-1}, x_i, x_{i+1}]$ and $[x_{j-1}, x_j, x_{j+1}]$ refer to entities $e_i$ and $e_j$ respectively. Tokens $x_{i-2}$ and $x_{i+2}$ are the preceding and succeeding tokens of the first mention. Note that a paragraph might include any number of mentions, each may contain any number of tokens. 

\subsection{ELMo Review}
\label{sec:elmorev}
For a given sequence, ELMo produces word representations on top of a 2-layer bi-RNN with character convolutions as input. For each direction, ELMo first computes a context-independent representation for each token at position $k$ by applying a character-based CNN. It then passes the token representations through a 2-layer LSTMs. As a result, each LSTM layer outputs a context-dependent representation $\overrightarrow{x}_{k,j}$ and $\overleftarrow{x}_{k,j}$ for layers $j \in \{1, \dots , L\}$ for the forward and backward directions respectively. The outputs of the last layer i.e.$\overrightarrow{x}_{k,L}$ and  $\overleftarrow{x}_{k,L}$ are given to a Softmax layer to predict the next and previous tokens $x_{k+1}$ and $x_{k-1}$ respectively. 

ELMo's objective is to jointly maximize the log likelihood of the forward and backward passes:
\vspace{-0.3cm}
\begin{align*}
    \mbox{\textit{ll}}_{\mbox{\tiny ELMo}} &= \sum_{k=1}^{n} \log p(x_k|\overrightarrow{x}_{k-1, L}, \overrightarrow{\Theta}_{\mbox{\tiny LM}}, \Theta_{\mbox{\tiny S}}, \Theta_{\mbox{\tiny x}})\\ &+ \sum_{k=1}^{n} \log p(x_{k}|\overleftarrow{x}_{k+1, L}, \overleftarrow{\Theta}_{\mbox{\tiny LM}}, \Theta_{\mbox{\tiny S}}, \Theta_{\mbox{\tiny x}})
\end{align*}
where $\overrightarrow{\Theta}_{\mbox{\tiny LM}}$, $\overleftarrow{\Theta}_{\mbox{\tiny LM}}$, $\Theta_{\mbox{\tiny S}}$ and $\Theta_{\mbox{\tiny x}}$ are the parameters for the forward and backward bi-RNNs, the Softmax and the char-CNN layers respectively.
\subsection{E-ELMo Model}
\label{sec:eelmorev}


As shown in Figure~\ref{fig:elmo}, E-ELMo is in fact ELMo with entities incorporated in the target layer, replacing the targets from the mention words to the grounded entity. In particular, the target for position $k\in I_i=\{i-2, i-1, i\}$ for the forward and $k \in J_i=\{i, i+1, i+2\}$ for the backward directions should be entity $e_i$. The log likelihood objective of the E-ELMo is the sum of the log likelihood for both words and entities as follows:
\begin{dmath}
  \mbox{\textit{ll}}_{\mbox{\tiny E-ELMo}} = \mbox{\textit{ll}}_{\tiny w} + \mbox{\textit{ll}}_{\tiny e} 
  \label{eq:join}
\end{dmath}
where
\begin{align*}
   \mbox{\textit{ll}}_{\tiny e} &=\sum_{i} (\sum_{k-1\in I_i}\log p(e_i|\overrightarrow{x}_{k-1, L}, \overrightarrow{\Theta}_{\mbox{\tiny LM}}, \Theta_{\mbox{\tiny E}}, \Theta_{\mbox{\tiny x}}) \\&+ \sum_{k+1\in J_i}\log p(e_i|\overleftarrow{x}_{k+1, L}, \overleftarrow{\Theta}_{\mbox{\tiny LM}}, \Theta_{\mbox{\tiny E}}, \Theta_{\mbox{\tiny x}}))\\
\end{align*}
where $\mbox{\textit{ll}}_{\tiny w}$ is equal to $\mbox{\textit{ll}}_{\tiny ELMo}$ minus the terms predicting entities and $\Theta_{\mbox{\tiny E}}$ is the entity parameters. Note that it is important to optimize the entity vectors on the unit sphere to yield qualitative embeddings for NED. To maximize Eq.~\ref{eq:join}, we consider several different configurations of E-ELMO:\\
\textbf{\normalsize Config-a:} We freeze all parameters except for $\Theta_{\mbox{\tiny E}}$.\\
\textbf{\normalsize Config-b:} All the parameters are fine-tuned.\\
\textbf{\normalsize Config-c:} All the parameters are fine tuned but we redefine Eq.~\ref{eq:join} to be  $\mbox{\textit{ll}}_{\mbox{\tiny E-ELMo}} = \mbox{\textit{ll}}_{\tiny e}$. 

\section{Local Entity Disambiguation Model}
To evaluate the effectiveness of the learned entity representations, we consider a simple local entity disambiguation model to rank the candidate entities.
Given a query mention $[x_{i-1}, x_{i}, x_{i+1}]$ with its local context $x_1, ... x_{i-2}, [x_{i-1}, x_i, x_{i+1}], x_{i+2}... x_{T}$, the context is fed to the E-ELMo corresponding to one of the config-a, b and c to produce $\overrightarrow{x}_{k,L}$ and $\overleftarrow{x}_{k,L}$ for all positions $k\in 1 \dots T$. The context representation for the mention is given by concatenating $\overrightarrow{f_c}=\frac{1}{|I_i|}\sum_{k\in I_i}(\overrightarrow{x}_{k, L})$ and$\overleftarrow{f_c} = \frac{1}{|I_j|}\sum_{k\in I_j}(\overleftarrow{x}_{k, L})$.

We also utilize the following basic features introduced by prior work \citep{l:yamada, l:yammada17, rel:ibm, rel:hoffman, l:phongle}. \normalsize Prior Compatibility ($f_p$): We consider $p(e|m)$, the prior probability that an entity $e$ is linked to a mention string $m$ as prior evidence. \normalsize String Matching ($f_s$): We use ten lexical features $f_s[1] \dots f_s[10]$ listed in \citep{l:shahbazh} to capture the lexical similarity between the query mention and the surface string of the entity. 
Our local model first transforms the scalar features $f_p, f_s[1] \dots f_s[10]$ through a bin layer~\citep{l:shahbazh, rel:ibm} to project each feature $f_.$ to a higher dimensional $\hat{f}_.$. The details of the binning and transformation can be found in Appendix A.1. The concatenated feature $[\hat{f}_p; \hat{f}_s[1] \dots \hat{f}_s[10]; \overrightarrow{f_c}; \overleftarrow{f_c};]$ and $\Theta_{E}[e]$, the representation for $e$ learned by E-ELMo, are then given to a 2-layer feed forward neural network with Relu activation to compute the final score for entity $e$.
\section{Experiments}
\subsection{Training E-ELMo} \label{sec:phi} We follow the same experimental setup as our baselines. We train E-ELMo on a subset $\varphi$ of Wikipedia corpus extracted by ~\citep{rel:hoffman}\footnote{Available at https://github.com/dalab/deep-ed \label{foot:hoff}} and also used by \cite{l:phongle}. We initialize$\overrightarrow{\Theta}_{\mbox{\tiny LM}}, \overleftarrow{\Theta}_{\mbox{\tiny LM}}, \Theta_{\mbox{\tiny S}},$ and $\Theta_{\mbox{\tiny x}}$ using the original ELMo pre-trained on 5\si{B} tokens. We initialize $\Theta_E$ for each entity to be the average of its title tokens in $\Theta_{s}$. The number of negative samples for both words and entitys is set to 8192. Training of E-ELMo is via AdaGrad \citep{l:duchi} with a learning rate 0.1 for 10 epochs.
\subsection{Training the Local Model} The binning layer in our local model projects a scalar $f_p$ to 15-\si{d} vector and each $f_s[k]$ to 10-\si{d} vector. The size of $\overrightarrow{f_c}$, $\overleftarrow{f_c}$, $\theta_e$ are each 512. We use the cross-entropy loss function and dropout of 0.7 on the feed forward neural network. Training is done by ADAM~\citep{l:adam} with learning rate 0.001.

\begin{table*}[!htbp]
\centering
\tiny
\begin{tabular}{l|c|c|c|c|c}
\hline models & MSB & AQ & ACE & CWEB & WW\\
\hline
global: \citep{l:guo} & 92 & 87 &88 & 77 & 84.5\\
global: \citep{rel:hoffman} & 93.7 $\pm$ 0.1  &   88.5 $\pm$ 0.4 & 88.5 $\pm$ 0.3 & 77.9 $\pm$ 0.1 & 77.5 $\pm$ 0.1\\
global: \citep{l:phongle} & 93.9 $\pm$ 0.2  & 88.3 $\pm$ 0.6 &89.9 $\pm$ 0.8 & 77.5 $\pm$ 0.1 & 78.0 $\pm$ 0.1\\
\hline
local: \hspace{0.1cm} $\mbox{E-ELMo}_b$ & 92.3 $\pm$ 0.1  & \textbf{90.1 $\pm$ 0.3} & 88.7 $\pm$ 0.1 & \textbf{78.4 $\pm$ 0.2} & 79.8 $\pm$ 0.2\\
local: \hspace{0.1cm} $\mbox{E-ELMo}_c$ & 92.0 $\pm$ 0.1  & 89.6 $\pm$ 0.1 & 87.6 $\pm$ 0.1 & 77.5 $\pm$ 0.3 & 78.4 $\pm$ 0.1\\

\hline
\end{tabular}\\
\caption{Results on five out-domain test sets}
\label{tab:wned}
\vspace{-0.5cm}
\end{table*}

\subsection{Entity Disambiguation Results} We first evaluate our NED systems on the two most commonly used benchmarks: AIDA-CoNLL \citep{l:hoffart} and TAC 2010 \citep{l:ji10}.  We follow the same setup as our baselines ~\cite{rel:hoffman, l:phongle} and use the same train, test, validation splits, the same candidate sets and the prior feature values $p(e|m)$. 
Please see Appendix A.2 for detailed information about the datasets used in the experiments. 
For AIDA-CoNLL, the existing literature has considered two different candidate sets: (aida-Yago), which is extracted and used by \citep{rel:hoffman, l:phongle} and (aida-HP): a less ambiguous one extracted by \citep{l:pershina}. In our experiments we evaluate our methods using both candidate sets.   


Table~\ref{tab:tac_aida} presents the results of our models as well as baseline local and global models on AIDA-CoNLL and TAC 2010. Our baselines include a large number of state-of-the-art local (seven) and global (eight) models proposed in recent years. In addition, we also consider an alternative NED system based on the original ELMo. In particular, this model, referred to as $\mbox{ELMO}_o$, is identical to our local NED model except that E-ELMo is replaced with the original ELMo and each entity $e$ is represented by averaging the sentence-level representations of all query mentions linking to $e$ in Wikipedia using the original ELMo. 



We first note that the proposed models are highly competitive, achieving substantial improvements over prior local methods, and even outperforming prior state-of-the-art global methods on both benchmarks. 
Comparing the performance of $\mbox{E-ELMo}_a$ with the other two variants, we observe that fine tuning the language model with the entity-aware objective tends to further improve the performance. 

We also note that the performance of the baseline model $\mbox{ELMo}_o$ is substantially lower on both datases, suggesting that the performance gain achieved by our models is not simply due to improved word representations from ELMo. Rather, it is critical to integrate entities into the language model to learn useful representation for NED. 
\begin{table}[t]
\centering
\tiny
\begin{tabular}{l|c|c|c}
\hline models & aida (HP) & aida (YAGO+KB) & tac (KB)\\
\hline
 local models &   &   &  \\
\hline
\citep{l:Francis} &  85.5 &  - & - \\
\citep{l:sil:16} &  86.2 &  - & 78.6 \\
\citep{l:yamada} &  90.9 &  87.2 & 84.6 \\
\citep{l:yammada17} & 94.7 & - & 87.7 \\
\citep{rel:hoffman} &  - &  88.8 & - \\
\citep{rel:ibm} & 94.0 & - & 87.4 \\
\citep{l:shahbazh} & 90.89 & - & 85.73\\
\hline
global models &  &  & \\
\hline
\citep{l:pershina} &  91.8 &  - & -\\
 \citep{l:chisholm} &  - &  88.7 & -\\
\citep{rel:google} &  92.7 &  91.0 & 87.2\\
\citep{l:yamada} &  93.1 &  91.5 & 85.5 \\
\citep{l:guo} &  - &  89.0 & -\\
\citep{rel:hoffman} &  - &  92.22 $\pm$ 0.14 & -\\
\citep{l:shahbazh} &  94.44 &  - & 87.9\\
\citep{l:phongle} &  - &  93.07 $\pm$ 0.27 & -\\
\hline
 this work (local) &   &   &  \\
\hline
\hspace{0.1cm}$\mbox{ELMo}_o$ (baseline) &  86.2 $\pm$ 0.16  & 84.01 $\pm$ 0.19  & 81.12 \\
\hspace{0.1cm}$\mbox{E-ELMo}_a$ &  95.41 $\pm$ 0.11 &  92.07 $\pm$ 0.16 & 88.15\\
\hspace{0.1cm}$\mbox{E-ELMo}_b$ &  96.22 $\pm$ 0.20 &  93.1 $\pm$ 0.22 & 87.36\\
\hspace{0.1cm}$\mbox{E-ELMo}_c$ &  \textbf{96.24 $\pm$ 0.12} &  \textbf{93.46 $\pm$ 0.14} & \textbf{88.27}\\
\hline
ablation (-prior -lexical features) &   &   &  \\
\hline
\hspace{0.1cm} $\mbox{E-ELMo}_b$  &- & 92.12  & 83.0  \\
\hspace{0.1cm} $\mbox{E-ELMo}_c$ &- & 92.30  & 84.4  \\
\hspace{0.1cm} \citep{rel:hoffman} - & -& 86.34 & 69.28  \\
\hline
\end{tabular}\\
\caption{Results on TAC 2010 and AIDA test-b}
\label{tab:tac_aida}
\vspace{-0.5cm}
\end{table}

The ablation rows in Table~\ref{tab:tac_aida} present the performance of our models versus the local attention based models ~\citep{rel:hoffman}  without using the prior and lexical features. As shown in the table our models see only slight reduction in performance when prior and lexical features are removed. In contrast, the local attention based model suffered a substantial performance dip. This suggests that the contextual dependencies captured by E-ELMo are significantly richer, possibly because E-ELMo not only captures syntactic and semantic dependencies but also  captures lexical and prior dependencies between the mention and the entity with the help of the char CNN in the first layer.

Table~\ref{tab:wned} presents the performance of $\mbox{E-ELMo}_b$ compared to state-of-the-art global models on five additional open domain datasets. Because these datasets are not widely used in prior studies, we can only compare to three prior methods, all of which are global methods. As can be seen from the table, $\mbox{E-ELMo}_b$ is very competitive compared to the prior state-of-the-art achieved by global models on these datasets as well. 

\subsection{Analysis of Results}
We take a closer look at the performance of our method in comparison with a popular global model \citep{rel:hoffman} (referred to as $\mbox{M}_g$) on the AIDA-CoNLL dataset (YAGO+KB), focusing on the 4400 test queries with gold in the candidate set. In particular we are interested in how entity frequencies and the number of entities in the same document influence the performance of our local model in comparison with the global model.  
\begin{table}[!htbp]
\centering
\tiny
\begin{tabular}{l|c|c|c|c}
\hline
Stats of entity $e$ & \# mentions & $\mbox{E-ELMo}_b$ & $\mbox{E-ELMo}_c$ & $\mbox{M}_g$ \\
\hline
frequency in Wikipedia &  &  \\
\hline
1-10 & 459 & 94.54 & 95.42 & 91.93\\
11-50 & 503 & 93.04 & 92.84 & 92.44 \\
$\ge$51 & 3438 & 94.50 & 96.24 & 94.21\\
\hline
\hline
\# of entities in doc &  &  \\
\hline
1-4 & 90 & 96.66 & 96.68 & 93.33\\
5-9 & 341 & 97.06 & 97.94 & 94.42\\
10-19 & 849 & 95.64 & 96.46 & 92.08\\
$\ge$20 & 3120 & 93.62 & 95.32 & 94.15\\
\hline
\end{tabular}\\
\caption{Comparing $\mbox{E-ELMo}_c$ and global $M_g$.}
\label{tab:cat}
\vspace{-0.5cm}
\end{table}
As shown in Table~\ref{tab:cat}, $\mbox{E-ELMo}_c$ has substantially higher performance compared to $M_g$ in dealing with entities with low frequency in the Wikipedia, suggesting that by unifying the entity representations with the word representation through E-ELMo, we can learn effective representations for rare entities. It is also observed that $\mbox{E-ELMo}_c$ performs consistently well on documents with different number of mentions, but the performance gap with $\mbox{m}_g$ is small for documents with over 20 mentions. This is consistent with expectation because such documents tend to benefit more from the global models.  


\section{Conclusions}
We introduced a novel approach for learning deep contextual entity representation by learning an entity aware extension of ELMo called E-ELMo. We also proposed a local entity disambiguation model which utilizes E-ELMo as its key component. The results demonstrate that our local model with very basic features achieves the best reported performance on AIDA-CoNLL and TAC-2010 with an improvement of about .5\% over the latest global model on AIDA-CoNLL. The model is also competitive to the global models on open domain datasets. 
\bibliography{emnlp-ijcnlp-2019}
\bibliographystyle{acl_natbib}
\section{Appendices}
\label{sec:appendix}

\subsection{Binning to project a scalar to a higher dimension}
\label{sec:bin}
Given a scalar variable $x$, binning projects $x$ to dimension $d$ as follows:
\begin{align*}
\textbf{p} = 
[e^{-(\epsilon_1 \parallel x - x_1 \parallel)^{2}} \dots e^{-(\epsilon_d \parallel x - x_d \parallel)^{2}}]
\end{align*}
Binning introduces parameters $\epsilon_i$ and $x_i$ for $i\in\{1 \dots d\}$ to project $x$ to vector $\textbf{p}$.

\subsection{Entity disambiguation datasts}
\label{sec:dsets}
We evaluate our NED on the following benchmarks: 
\begin{itemize}
    \item AIDA-CoNLL \citep{l:hoffart}: This dataset which is one of the biggest dataset for NED, contains training (AIDA-train),
validation (AIDA-A) and test (AIDA-B) sets. \item MSNBC (MSB), AQUAINT (AQ) and
ACE2004 (ACE) datasets \citep{l:guo}
\item WNED-WIKI (WW) and WNED-CWEB
(CWEB): These datasets are bigger and are built from the
ClueWeb and Wikipedia corpora by \cite{l:guo, l:gabrilovich} 
\item  TAC 2010 \citep{l:hoffart} and TAC 2010 \citep{l:ji10}: This dataset is very popular with the baselines for both local and global model. 

To follow similar setup to our baselines ~\cite{rel:hoffman, l:phongle} we use the train, test, validation splits and also candidate sets and also prior values $p(e|m)$ for all the the datasets from ~\cite{rel:hoffman}\footnote{Available at https://github.com/dalab/deep-ed \label{foot:hoff}}
\end{itemize}
Table~\ref{tab:dsets} shows the statistics of the datasets:
\begin{table}[!htbp]
\tiny
\centering
\begin{tabular}{l|c|c|c|c}
\hline Dataset & mentions & docs & mention per doc & candidate gen recall\\
\hline
AIDA-train & 18448 & 946 & 19.5 & -\\
\hline
AIDA-A (valid) & 4791 & 216 & 22.1 & 96.9\%\\
\hline
AIDA-B (test) & 4485 & 231 & 19.4 & 98.2\%\\
\hline
MSNBC & 656 & 20 & 32.8 & 98.5\%\\
\hline
AQUAINT & 727 & 50 & 14.5 & 94.2\%\\
\hline
ACE2004 & 257 & 36 & 7.1 & 90.6\%\\
\hline
WNED-CWEB & 11154 & 320 & 34.8 & 91.1\%\\
\hline
WNED-WIKI & 6821 & 320 & 21.3 & 92\%\\
\hline
TAC-2010(test) & 1020 & 1013 & 1 & 93\%\\
\hline
\end{tabular}\\
\caption{statistics about the datasets used in our NED}
\label{tab:dsets}
\end{table}
 
\end{document}